# Credit Card Fraud Detection

**Iva Popova**
ETIT-KIT, Germany

**Hamza A. A. Gardi**
IIIT at ETIT-KIT, Germany

**Abstract**

Credit card fraud remains a significant challenge due to class imbalance and fraudsters mimicking legitimate behavior. This study evaluates five machine learning models - Logistic Regression, Random Forest, XGBoost, K-Nearest Neighbors (KNN), and Multi-Layer Perceptron (MLP) on a real-world dataset using undersampling, SMOTE, and a hybrid approach. Our models are evaluated on the original imbalanced test set to better reflect real-world performance. Results show that the hybrid method achieves the best balance between recall and precision, especially improving MLP and KNN performance.

## 1. Introduction

Financial fraud is a significant issue that has been continuously increasing over the past few years due to the ever-growing volume of online transactions conducted with credit cards. Credit card fraud (CCF) refers to a type of fraud in which an individual other than the cardholder unlawfully conducts transactions using a card that is stolen, lost, or otherwise misused [1]. CCF has resulted in billions of dollars in losses for banks and other online payment platforms. According to the Federal Trade Commission (FTC), there were 449,076 reports of CCF in 2024, representing a 7.8% increase from the previous year [2]. Given this trend, new methods must be employed to capture patterns and dependencies in the data. As a result, machine learning techniques have increasingly been used in automated fraud detection systems. These models are trained to distinguish between fraudulent and non-fraudulent transactions. However, a significant challenge in this training process is the highly imbalanced nature of the dataset - the number of fraudulent transactions is negligible compared to the total number of transactions. Consequently, it becomes difficult for the model to effectively learn from the minority class, limiting its ability to accurately detect fraud.

This paper aims to evaluate and compare the effectiveness of different resampling strategies - undersampling, SMOTE oversampling, and a hybrid of both for addressing class imbalance in credit card fraud detection. It focuses on realistic evaluation by testing all models on the original, imbalanced dataset and compares the performance of five machine learning models: Logistic Regression, Random Forest, XGBoost, K-Nearest Neighbors (KNN), and a Multi-Layer Perceptron (MLP).

## 2. Related Work

Credit card fraud detection is highly challenging due to the extremely low occurrence of fraudulent transactions compared to legitimate ones. One of the main difficulties is that fraudulent transactions often mimic normal behavior, making them hard to detect. As fraudsters constantly adapt to bypass detection systems, it becomes even more important for fraud detection models to distinguish between the two classes effectively.

To address the issue of class imbalance, various resampling techniques have been explored. Two of the most common ones are random undersampling of the majority class and oversampling of the minority class. For example, [3] tested various fraud-to-legitimate transaction ratios, such as 50:50, 10:90, and 1:99 - and found that a 10:90 distribution provided the best results. This ratio effectively balanced the detection of fraudulent cases while maintaining a realistic dataset. The experiments were conducted using Naïve Bayes, k-NN, and Logistic Regression models. Similarly, Sahin et al. [4] used a combination of stratified sampling and cost-sensitive decision trees, along with SMOTE to oversample fraud cases, which helped increase the model's ability to detect fraud without losing too much useful data.

Earlier studies, such as those by Stolfo et al. and Fan et al. [5], explored techniques like meta-learning and ensembles to handle class imbalance. They emphasized the importance of not just improving model accuracy, but also making sure the model remains sensitive to detecting rare fraud cases. More recent research has shown that hybrid techniques, which combine oversampling and undersampling - can produce even better results. One such study – [6], found that this approach helps improve both recall and F1-score.

However, one issue in many of these studies is that models are often tested on balanced test sets, which don't reflect real-life conditions where fraud is still extremely rare.

Our study tackles this issue - instead of using a resampled test set, we evaluate all our models on the original imbalanced test data. This gives us a more accurate view of how well the models would perform in a real-world fraud detection setting.

## 3. Methodology

### 3.1 Dataset

The most widely used dataset for credit card financial fraud detection is the one provided by the ULB Machine Learning Group, available for download on Kaggle [7]. It contains transactions made over two days in September 2013 by European cardholders and includes 31 numerical features. For privacy and security reasons, 28 out of the 31 features are anonymized using Principal Component Analysis (PCA). Only three features are directly interpretable: "Time," "Amount" (the transaction amount), and "Class" (where 0 indicates a non-fraudulent transaction and 1 indicates a fraudulent one). The dataset is highly imbalanced, with just 492 out of 284,807 transactions being fraudulent (only 0.173%) [1].

### 3.2 Feature Extraction

The Credit Card Fraud Dataset has already undergone feature extraction using PCA for the anonymized features (V1 to V28). In our implementation, we additionally scale the "Amount" and "Time" features to reduce the impact of outliers.

### 3.3 Preprocessing

Due to the highly imbalanced class distribution, preprocessing is an essential part of achieving good model performance. Machine learning algorithms often fail to capture meaningful patterns when the classes are not evenly represented. In our work, we examine three different methods for balancing the two classes.

#### 3.3.1 Undersampling

Undersampling is a technique that reduces the number of instances in the majority class by randomly removing samples until the desired class ratio is achieved. In our case, the number of non-fraudulent transactions is reduced to 492 to match the number of fraudulent ones. While this results in a significantly smaller training set (984 instances in total) and faster training times, it also leads to the loss of valuable information from discarded legitimate transactions.

#### 3.3.2 SMOTE Oversampling

SMOTE (Synthetic Minority Over-sampling Technique) is a technique that generates synthetic instances of the minority class to match the size of the majority class. Unlike simple duplication, SMOTE creates new samples by interpolating between an existing minority instance and one of its k-nearest neighbors. This enriches the minority class and helps the model generalize better. However, SMOTE has its drawbacks, including increased computational time and the potential for class overlap, where synthetic points may fall into the majority class space, confusing the classifier.

#### 3.3.3 Hybrid Sampling Techniques

To achieve high performance while maintaining control over the ratio between non-fraudulent and fraudulent data, we explore a hybrid sampling technique [6]. This method combines undersampling of the majority class with SMOTE-based oversampling of the minority class. The key advantage of this approach is its flexibility - it allows for adjustment of the class ratio. Through practical experimentation (detailed in the next section), we discovered that increasing the proportion of fraudulent transactions in the training data to more around 2,5 % - while still maintaining a higher number of non-fraudulent samples - helps reduce overfitting and improves overall model performance.

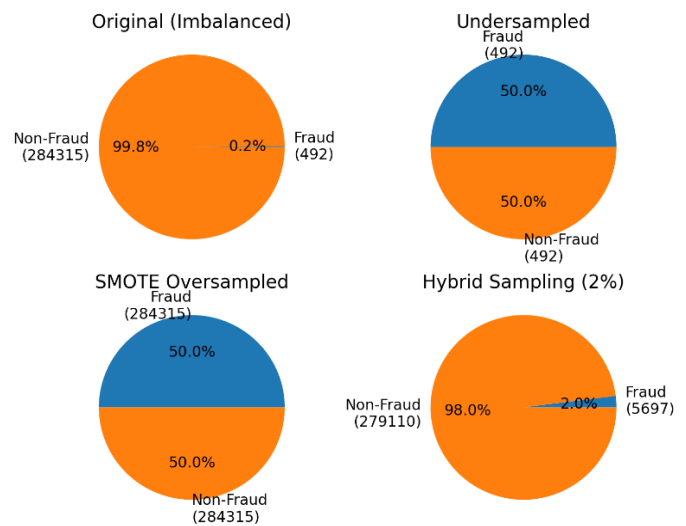

*Fig. 1: Class distributions for different sampling techniques: Original (Imbalanced), Undersampled, SMOTE Oversampled, Hybrid Sampling (2%)*

## 3.4 Metrics
### 3.4.1 Accuracy

In most machine learning tasks, accuracy is a commonly used metric that reflects the proportion of correctly classified instances over the total number of instances. It provides a general idea of how close the model's predictions are to the actual values:

$$Accuracy = \frac{TP + TN}{TP + TN + FP + FN} \quad (1)$$

However, in the context of credit card fraud detection - where the dataset is highly imbalanced - accuracy can be misleading. A model may achieve very high accuracy simply by correctly predicting the abundant non-fraudulent transactions, while still failing to detect fraudulent ones. Therefore, additional metrics such as precision, recall, and F1-score are essential for evaluating the model's true effectiveness.

### 3.4.2 Precision

Precision measures the proportion of correctly predicted positive cases (true positives) to all cases predicted as positive (true positives + false positives). It reflects how reliable the model's positive predictions are:

$$Precision = \frac{TP}{TP + FP} \quad (2)$$

High precision indicates that when the model predicts fraud, it is usually correct - minimizing false alarms.

### 3.4.3 Recall

Recall, also known as sensitivity or the true positive rate, measures the proportion of actual fraud cases that were correctly identified by the model. It is defined as:

$$Recall = \frac{TP}{TP + FN} \quad (3)$$

A recall of 1 means that the model identified all actual fraud cases, with no false negatives. In fraud detection, maximizing recall is especially important, as failing to detect fraudulent transactions can have serious consequences.

## 3.5 Models

In machine learning, classification refers to a type of prediction task where the goal is to assign input data (also known as independent variables) to one or more predefined categories. These categories can range from just two to many, depending on the problem. Classification techniques can be applied to both structured and unstructured data. A "classifier" is an algorithm designed to categorize new data into one of the defined classes. There are three main types of classification problems: binary classification, multi-class classification, and multi-label classification. The detection of credit card fraud detection is a binary classification problem. In this work we use the following classifiers for it: [8]

### 3.5.1 Logistic Regression

Logistic Regression is a supervised learning algorithm often used for classification problems. It predicts the value of a target categorical or numerical variable using input features, which can be either continuous or discrete. The algorithm is capable of handling both types of data simultaneously, making it flexible and widely used in machine learning. Its core function is to estimate the probability that a given input belongs to a specific class, making it particularly useful for binary and multi-class classification problems. [8]

The logistic regression model calculates the probability that a given input x belongs to class 1 using the sigmoid function:

$$P(y = 1 \mid x) = \frac{1}{1+e^{-(\beta_0 + \beta_1 x_1 + \beta_2 x_2 + \cdots + \beta_n x_n)}} \quad (4)$$

Where:

- $P(y=1|x)$: Probability that the output y is 1 given inputs x
- $\beta_0$: Intercept (bias)
- $\beta_1, \beta_2, \ldots, \beta_n$: Coefficients of the features $x_1, x_2, \ldots, x_n$
- e: Euler's number (approx. 2.718)

This formula maps the linear combination of input features to a value between 0 and 1, representing the predicted probability. If this probability is greater than 0.5 (by default), the model predicts class 1; otherwise, class 0.

### 3.5.2 Random Forest

Random Forest (RF) is an ensemble learning method that combines multiple decision trees to improve performance in both classification and regression tasks. It is particularly effective at capturing complex, non-linear relationships within the data. By aggregating the predictions of many individual trees, Random Forest reduces variance and mitigates the risk of overfitting.

Given a training dataset P = (p₁, ..., pₙ) with corresponding labels Q = (q₁, ..., qₙ), the Random Forest algorithm employs a technique called bagging (bootstrap aggregating). This involves drawing random samples with replacement from the training data X times to build a series of independent decision trees. Each tree is trained on a different bootstrap sample.

The final prediction for an input $\dot{R}$ is made by averaging the predictions of all individual trees:

$$\hat{y} = \frac{1}{X} \sum_{x=1}^{X} f_x(\dot{R}) \quad (5)$$

Where $f_x(\dot{R})$ is the prediction of the x-th tree for the input $\dot{R}$. [1]

### 3.5.3 XGBoost

XGBoost is a highly effective machine learning algorithm that can be applied to almost any classification task. Its power lies in combining gradient boosting with regularization techniques, which not only improve accuracy but also help prevent overfitting. XGBoost is also known for its ability to handle missing values, support parallel computation, and efficiently process large datasets, making it a flexible and reliable option for various machine learning applications. [1]

Boosting is built on the idea that combining multiple weak (low-accuracy) classifiers can create a single strong (high-accuracy) model. Gradient Boosting Machines (GBMs) are a specific type of boosting algorithm that work by adding one weak learner at a time. At each step, the next learner is chosen based on the gradient (or direction) of the loss function, aiming to reduce the overall error of the model. [9]

### 3.5.4 K-Nearest Neighbours (KNN)

The K-Nearest Neighbours (KNN) algorithm is a supervised machine learning method used for both classification (discrete labels) and regression (continuous labels) tasks. It assigns a class or value to a new data point by comparing it to existing training data. To make predictions, KNN measures the similarity between the input data and the stored training examples, since it retains all training data during the process [10]. [11]

In this study, a k value of 5 nearest neighbors was chosen to maintain a good balance between model complexity and generalization. The algorithm typically uses Euclidean distance to measure how close each training sample is to the input query, selecting the k samples with the shortest distances. [12]

### 3.5.5. Multilayer Perceptron (MLP)

The Multilayer Perceptron (MLP) implemented in this model is a feedforward artificial neural network composed of three layers: an input layer, two hidden layers, and an output layer. Each neuron in the network applies an activation function that calculates the weighted sum of its inputs along with a bias term. This structure enables the network to learn meaningful patterns and relationships within the data. In this case, the MLP uses two hidden layers with 32 and 16 units, respectively, both employing the ReLU activation function to introduce non-linearity and improve learning efficiency. The output layer consists of a single neuron with a sigmoid activation function, making it suitable for binary classification tasks, such as detecting credit card fraud. The model was trained using the backpropagation algorithm, a standard and effective method for supervised learning in neural networks. This involves a forward pass to generate predictions, followed by a backward pass where the model computes gradients of the loss with respect to its weights and updates them accordingly to minimize error.

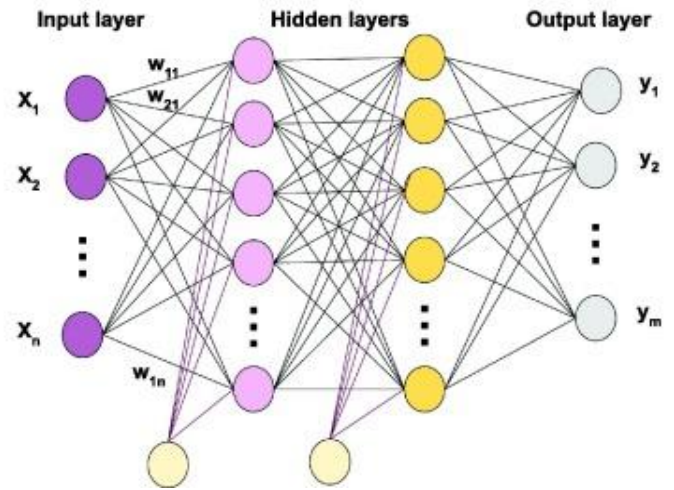

*Fig. 2: Structure of a simple MLP (Multi-Layer Perceptron)*

The model is optimized using Adam, a stochastic gradient-based optimizer known for its efficient weight updates and faster convergence. For training the dataset was split into training and testing sets with a 75/25 ratio while preserving class distribution using stratification. Further, the training set was split again to create a validation set (15% of the training data), ensuring reliable performance tracking during training. The model was then trained over 15 epochs with a batch size of 256, using both the training and validation sets to monitor learning progress. Although we experimented with deeper network architectures, this configuration was chosen as it provided a balanced trade-off between model complexity and computational efficiency.

MLPs work well for fraud detection because they can learn complex and hidden patterns in transaction data. They can also adjust over time to new types of fraud by updating their weights during training. However, compared to some other machine learning methods, they usually need more data and take more time and computing power to train. [13]

## 4. Performance Evaluation and Results
### 4.1 Imbalanced Dataset

First, we trained our models on the highly imbalanced dataset - 0.173% fraudulent transactions and the remainder non-fraudulent. The dataset was split into training and testing sets with a 75:25 ratio, and the proportion of fraud to non-fraud transactions remained consistent across both splits. To ensure reproducibility and fair comparison, all models were trained and evaluated using the same random seed.

A detailed evaluation of five machine learning models - Logistic Regression, Random Forest, XGBoost, K-Nearest Neighbors (KNN), and Multilayer Perceptron (MLP) - was conducted using precision, recall, F1-score, and accuracy as key performance metrics. While all models in this study achieved extremely high accuracy scores (≥ 99.9%), this metric is misleading in the context of credit card fraud detection, where the dataset is heavily imbalanced. In such datasets, non-fraudulent transactions (class 0) vastly outnumber fraudulent ones (class 1). As a result, models become biased toward predicting non-fraud, and in most cases, they would be correct. Consequently, all models achieved high accuracy, precision, recall, and F1-scores for class 0.

Due to this imbalance, a model could achieve high accuracy simply by predicting every transaction as non-fraudulent (i.e., always outputting class 0). However, a model that fails to detect fraud may be "accurate" on paper, yet it provides little real-world value in terms of security or risk mitigation. Therefore, in the remainder of our experiments, we focus only on the metrics for class 1 (fraudulent transactions), as these differ significantly across experimental setups. Particular emphasis is placed on recall and F1-score, which better reflect a model's ability to detect fraudulent transactions. The five models are evaluated on the original imbalanced test set and their performance is summarized as follows:

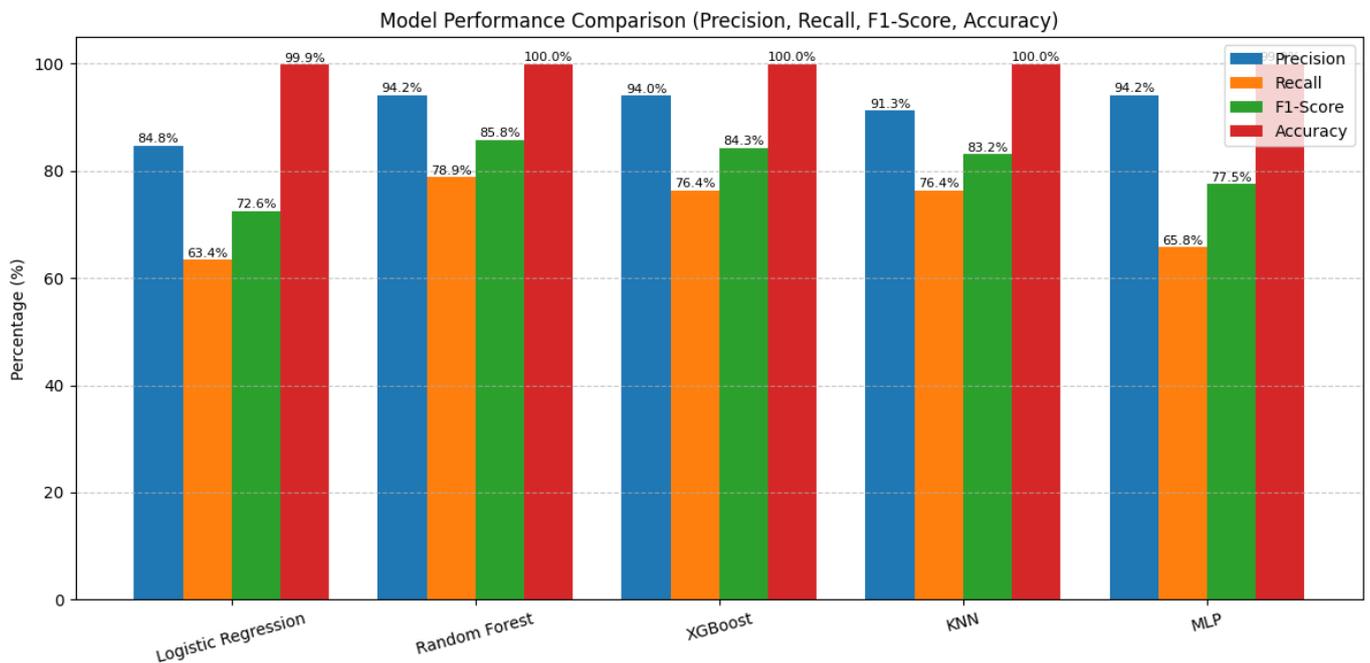

*Fig. 3: Comparison of model performance across four metrics (Precision, Recall, F1-Score, and Accuracy) on the imbalanced dataset*

### 4.1.1 Logistic Regression

Logistic Regression yielded the weakest performance among the evaluated models, with 84.8% precision, 63.4% recall, and an F1-score of 72.6%. Despite maintaining a high accuracy rate, it fails to detect a significant number of fraudulent transactions. This limits its suitability for high-risk applications like credit card fraud detection, where missing fraud is more costly than raising a few false alarms.

### 4.1.2 Random Forest

Random Forest demonstrated the most robust performance overall. It achieved a precision of 94.2%, a recall of 78.9%, and the highest F1-score of 85.8%. This indicates that the model is both highly effective at identifying fraud cases (high recall) and minimizes false positives (high precision). The strong F1-score reflects a balanced performance, making Random Forest the most reliable model for fraud detection on the imbalanced dataset.

### 4.1.3 XGBoost

XGBoost closely followed Random Forest with 94.0% precision, 76.4% recall, and an F1-score of 84.3%. While its recall is slightly lower than that of Random Forest, XGBoost still performs strongly across all metrics. Given its efficiency and scalability, it remains a highly competitive alternative for fraud detection tasks.

### 4.1.4 K-Nearest Neighbors (KNN)

KNN showed solid performance with 91.3% precision, 76.4% recall, and an F1-score of 83.2%. Although it falls slightly behind tree-based methods in terms of precision, its recall is comparable to that of XGBoost. This suggests that KNN can effectively detect fraud, though it may produce more false positives.

### 4.1.5 Multilayer Perceptron (MLP)

The MLP model achieved the highest precision (94.2%), indicating a very low false positive rate. However, its recall was lower at 65.8%, resulting in an F1-score of 77.5%. While MLP is good at correctly identifying legitimate transactions, it misses a larger portion of actual fraud cases compared to tree-based models. This trade-off suggests the need for further tuning or a deeper architecture to improve its fraud detection capability.

In the next set of experiments, we will alter the class distribution of the dataset to achieve a more reasonable balance between precision and recall, while maintaining a high F1-score. Recall is essential, as it measures how many actual fraudulent transactions were correctly detected. At the same time, we also aim to maintain a high precision, since false positives can lead to legitimate transactions being flagged, causing inconvenience to users who then need to verify their activities.

### 4.2 Undersampling

We first applied the dataset balancing technique of undersampling. A unique subset of non-fraudulent transactions, equal in number to the fraud cases, was randomly selected to create a 1:1 balance between fraud and non-fraud samples.

| Subset | Class 0 (Non-Fraud) | Class 1 (Fraud) |
|---|---|---|
| Original Full | 284315 | 492 |
| Original Train | 213236 | 369 |
| Original Test | 71079 | 123 |
| Undersampled Full | 492 | 492 |
| Undersampled Train | 372 | 366 |
| Undersampled Test | 120 | 126 |

*Table 1: Class distribution in the original and undersampled datasets, including training and test splits.*

The figure above illustrates the class distribution in the dataset. The original highly imbalanced data was undersampled to contain 492 non-fraudulent and 492 fraudulent transactions. The dataset was then split into training and testing sets (75:25 ratio). The balanced undersampled test set contained 120 non-fraudulent and 126 fraudulent transactions. All models were trained on the undersampled training set, where the class distribution was nearly balanced, ensuring that the model learned fairly from both classes. The models were then evaluated on both the balanced test set and the original imbalanced test set to assess their generalization and applicability in real-world scenarios.

The results for class 1 (fraudulent transactions) from the balanced test set are shown in the table below:

| Model | Precision | Recall | F1-Score | Accuracy |
|---|---|---|---|---|
| Logistic Regression | 0,9225 | 0,9917 | 0,9558 | 0,9553 |
| Random Forest | 0,9084 | 0,9917 | 0,9482 | 0,9472 |
| XGBoost | 0,9147 | 0,9833 | 0,9478 | 0,9472 |
| KNN | 0,8855 | 0,9667 | 0,9243 | 0,9228 |
| MLP | 0,8879 | 0,8374 | 0,8619 | 0,8659 |

*Table 2: Model performance (Precision, Recall, F1-Score, Accuracy) evaluated on the balanced undersampled test set.*

The comparison of model performance on the undersampled dataset (top table) versus the original imbalanced dataset (bottom table) reveals a classic trade-off influenced by data balancing. On the undersampled dataset, all models achieved very high recall (ranging from 0.8374 to 0.9917), and precision remained above 0.88 for all models. Logistic Regression stood out with the best overall balance, achieving the highest F1-score (0.9558) and accuracy (0.9553), followed closely by Random Forest and XGBoost. This indicates that when trained and tested on a balanced dataset, models are effective at identifying both fraud and non-fraud cases.

However, in real-world scenarios, data is not evenly distributed between classes. Therefore, we must ensure that our models generalize well when trained on balanced undersampled data and tested on the real-world, original

test set. The results from the evaluation using the original imbalanced test data are presented below:

| Model | Precision | Recall | F1-Score | Accuracy |
|---|---|---|---|---|
| Logistic Regression | 0,0501 | 0,9024 | 0,0949 | 0,9703 |
| Random Forest | 0,0831 | 0,9593 | 0,1529 | 0,9816 |
| XGBoost | 0,034 | 0,9106 | 0,0656 | 0,9552 |
| KNN | 0,0499 | 0,8455 | 0,0942 | 0,9719 |
| MLP | 0,0111 | 0,8943 | 0,0219 | 0,8622 |

Table 3: Model performance (Precision, Recall, F1-Score, Accuracy) evaluated on the original imbalanced test set after training on the undersampled dataset.

When these same models are evaluated on the original imbalanced test set, precision drops drastically for all models (e.g., 0.0501 for Logistic Regression, 0.0831 for Random Forest), while recall remains relatively high. This sharp decline in precision highlights a critical issue: the models produce many false positives when fraud is rare in the data. Random Forest achieved the best balance in this context, with the highest F1-score (0.1529) and recall (0.9593), suggesting it handles imbalanced distributions more robustly than others. These results demonstrate the importance of evaluating models on imbalanced data to avoid misleadingly optimistic performance on artificially balanced subsets.

### 4.3 Oversampling with SMOTE

In the next experiment, SMOTE (Synthetic Minority Over-sampling Technique) was used to balance the training data by synthetically creating additional instances of the minority class to match the number of majority class samples. First, we split the data into training and testing sets (75:25), keeping the testing set imbalanced to avoid data leakage. Then, SMOTE was applied only to the training set to generate synthetic fraud samples, helping the models learn from a balanced class distribution. As before, the models were evaluated on both the balanced SMOTE test set and the original imbalanced test set.

The class distributions for all data subsets are shown below:

| Subset | Class 0 (Non-Fraud) | Class 1 (Fraud) |
|---|---|---|
| Original Full | 284315 | 492 |
| Original Train | 213236 | 369 |
| Original Test | 71079 | 123 |
| SMOTE Full | 284315 | 284315 |
| SMOTE Train | 213236 | 213236 |
| SMOTE Test | 71079 | 71079 |

Table 4: Class distribution in the original and oversampled with SMOTE datasets, including training and test splits.

The results from the balanced SMOTE test dataset for class 1 show idealized model performance, with all models achieving both high precision and recall. XGBoost stands out with the best overall F1-score (0.9374) and accuracy (0.9406), closely followed by Logistic Regression and Random Forest.

| Model | Precision | Recall | F1-Score | Accuracy |
|---|---|---|---|---|
| Logistic Regression | 0,9739 | 0,8834 | 0,9264 | 0,9298 |
| Random Forest | 0,9975 | 0,8612 | 0,9243 | 0,9295 |
| XGBoost | 0,9916 | 0,8888 | 0,9374 | 0,9406 |
| KNN | 0,997 | 0,8463 | 0,9155 | 0,9219 |
| MLP | 0,9989 | 0,8531 | 0,9202 | 0,9261 |

Table 5: Model performance (Precision, Recall, F1-Score, Accuracy) evaluated on the balanced oversampled with SMOTE test set.

In contrast, when evaluated on the original imbalanced test set, all models demonstrate high recall, indicating they are effective at identifying fraud cases. However, precision drops significantly, especially for Logistic Regression (0.0608) and XGBoost (0.1649), reflecting a high number of false positives. Despite the high overall accuracy values (above 97%), these figures are misleading due to the overwhelming majority of non-fraud cases. MLP and Random Forest strike the best balance under this condition, with MLP achieving the highest F1-score (0.699).

| Model | Precision | Recall | F1-Score | Accuracy |
|---|---|---|---|---|
| Logistic Regression | 0,0608 | 0,8862 | 0,1138 | 0,9762 |
| Random Forest | 0,4023 | 0,8374 | 0,5435 | 0,9976 |
| XGBoost | 0,1649 | 0,8618 | 0,2768 | 0,9922 |
| KNN | 0,3561 | 0,8049 | 0,4938 | 0,9971 |
| MLP | 0,6084 | 0,8211 | 0,699 | 0,9988 |

Table 3: Model performance (Precision, Recall, F1-Score, Accuracy) evaluated on the original imbalanced test set after training on the oversampled with SMOTE dataset.

This highlights that while performance on synthetically balanced data can appear impressive, real-world deployment requires careful consideration of precision. Overall, MLP and Random Forest emerge as the most promising models, demonstrating the most consistent performance across both evaluation scenarios.

### 4.4 Hybrid Sampling (SMOTE + Undersampling)

Our final experiment demonstrated that training the models on balanced data does not guarantee high-quality solutions on the original testing dataset, where the classes are not evenly distributed. Consequently, our objective in this last experiment is to identify the optimal ratio for the imbalanced data between non-fraudulent and fraudulent instances to train our models, thereby achieving higher recall, which is essential for financial fraud detection, while also maintaining relatively high precision to avoid unnecessarily stressing users to authenticate themselves due to false positives. Therefore, we must determine the ratio that strikes a balance between recall and precision, while maintaining a high F1-score.

This hybrid method combines SMOTE oversampling and random undersampling to achieve a controlled class ratio in the training set while preserving the natural imbalance in the test set for realistic evaluation. The process begins with loading and preprocessing the dataset. The dataset is then divided into training and test sets before any resampling is applied, which is crucial to prevent information leakage from synthetic data into the evaluation phase. The test set remains untouched throughout the experiment to ensure fair evaluation across all model configurations.

A range of fraud ratios (from 1% to 50%) is defined for experimentation. For each ratio, the training set is adjusted in two steps: first, SMOTE is applied to oversample the minority class (fraud) to the desired count. Then, random undersampling is employed to reduce the number of non-fraud samples, maintaining the target fraud ratio. We plot the precision and recall for each of the ratios and select a ratio that provides both high precision and high recall on the original testing set. The plot below illustrates the results for the Logistic Regression model:

The optimal ratio for training the models often involves increasing the fraudulent data from 0.173% (original dataset) to 1% - 2.5%, which allows the model to observe more fraudulent examples while still preserving a significant data imbalance. After selecting the best ratio for each model, we train it with the new training data and evaluate the trained model on the original, unchanged testing dataset.

The comparative analysis of baseline and hybrid sampling strategies demonstrates that the hybrid approach significantly enhances the models' ability to detect fraud by correcting the severe class imbalance in the training data. When evaluated on the original, imbalanced test set, all models exhibit notable improvements in recall and balanced F1-scores, making hybrid sampling a robust preprocessing method for real-world fraud detection tasks.

#### 4.4.1 Logistic Regression (LR)

Beginning with Logistic Regression (LR), the recall improves from 0.634 to 0.781 (+23.2%), a critical shift given that the model initially failed to identify 45 out of 123 frauds. With hybrid training, only 27 were missed. This change increased the F1-score from 0.726 to 0.762 (+5%), despite a slight decline in precision from 0.848 to 0.744. This trade-off is expected: more frauds are detected, but at the cost of a few additional false positives.

#### Random Forest

Random Forest already performs well on imbalanced data but still benefits from the hybrid method. The recall rises from 0.789 to 0.837 (+6.1%), correctly detecting 103 frauds instead of 97. The F1-score improves from 0.858 to 0.862, while precision drops modestly from 0.942 to 0.888. These results reflect a better balance between sensitivity and specificity with minimal compromise in overall performance.

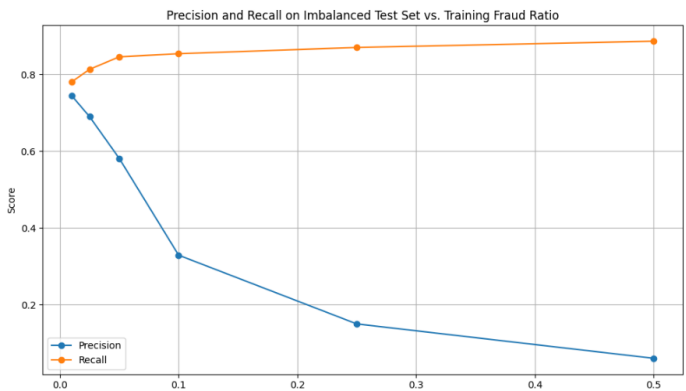

*Fig. 4: Precision and Recall on Imbalanced Test Set vs. Training Fraud Ratio*

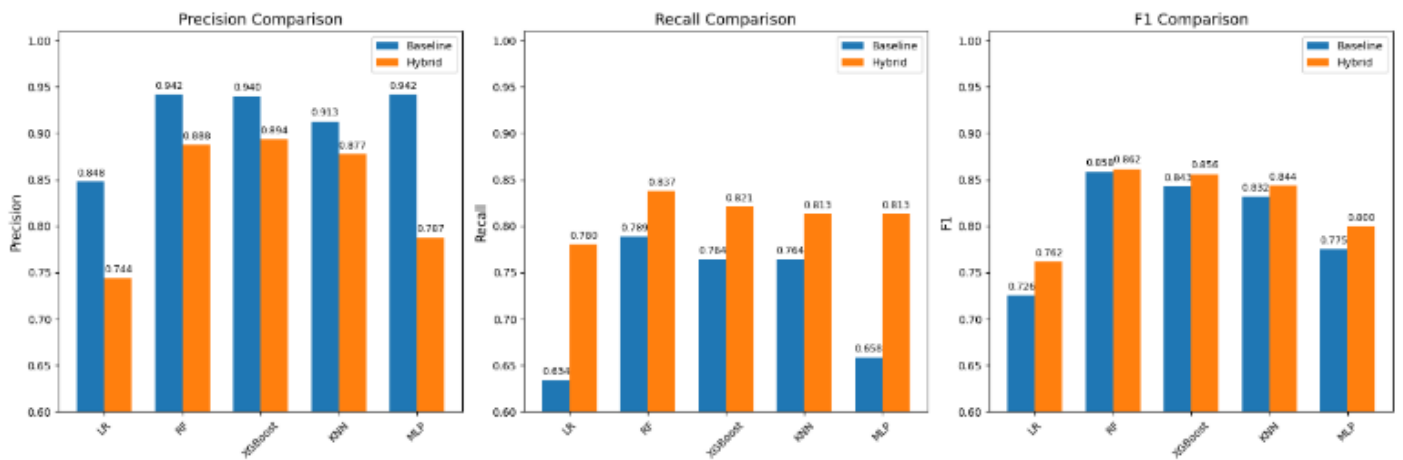

*Fig. 5: Comparison of Precision, Recall, and F1-Score Between Baseline and Hybrid Sampling Across All Models*

### 4.4.2 XGBoost

XGBoost exhibits a similar pattern: recall improves from 0.764 to 0.821 (+7.5%), with 101 out of 123 frauds detected compared to 94 in the baseline. Precision decreases slightly from 0.940 to 0.894, and the F1-score improves from 0.843 to 0.856 (+1.5%). These results indicate that even models that already generalize well benefit from hybrid class balancing during training.

### 4.4.3 K-Nearest Neighbors (KNN)

For K-Nearest Neighbors (KNN), the recall increases from 0.764 to 0.813 (+6.4%), and the F1-score rises from 0.832 to 0.844 (+1.4%). Precision decreases slightly from 0.913 to 0.877. Given KNN's vulnerability to class imbalance (due to neighbor-based majority voting), this result supports the effectiveness of balancing the training set.

### 4.4.4 Multi-Layer Perceptron (MLP)

The Multi-Layer Perceptron (MLP) benefits the most from hybrid sampling. Recall jumps from 0.659 to 0.813 (+23.4%), demonstrating its strong dependence on balanced data. Precision drops from 0.942 to 0.787 (−16.4%), but the F1-score improves from 0.775 to 0.800 (+3.2%) - this trade-off is justified in scenarios where accurately detecting fraudulent behavior takes priority over minimizing false positives. The model misses only 23 frauds instead of 45 - nearly halving the number of undetected fraud cases.

## 5. Conclusion

This study explored and evaluated various sampling strategies to tackle the pronounced class imbalance in credit card fraud detection, comparing baseline performance with three primary techniques: undersampling, SMOTE-based oversampling and a hybrid method combining SMOTE with undersampling. Through extensive experiments using multiple classifiers - Logistic Regression, Random Forest, XGBoost, K-Nearest Neighbors, and Multi-Layer Perceptron - we demonstrated that resampling methods such as hybrid approach, can significantly enhance the detection of rare fraudulent transactions.

Among the key findings, recall consistently improved across all models with hybrid sampling, with increases of up to 23% in some cases. This reflects a crucial gain in the model's ability to correctly identify fraud, which is paramount in real-world financial systems where the cost of undetected fraud is high. While precision slightly declined due to more false positives, the rise in F1-score for all models shows that the overall classification performance became more balanced and effective. Notably, models such as Logistic Regression and MLP, which are more sensitive to class distribution, benefited the most from the hybrid technique, whereas tree-based models like Random Forest and XGBoost also showed measurable, though more modest, improvements, as they showed strong performance even on the imbalanced original dataset.

Ultimately, our results reinforce that class imbalance cannot be ignored in fraud detection, and that resampling - particularly hybrid resampling - should be an integral part of the model development pipeline. By improving the model's sensitivity to fraudulent activity without drastically sacrificing overall accuracy, hybrid sampling offers a practical and scalable solution for real-world financial fraud prevention systems.